\documentclass[sigconf, language=english, natbib=true, screen,nonacm,review=false,timestamp=false]{acmart}

\copyrightyear{2023}
\acmYear{2023}
\setcopyright{acmlicensed}\acmConference[CIKM '23]{Proceedings of the 32nd ACM International Conference on Information and Knowledge Management}{October 21--25, 2023}{Birmingham, United Kingdom}
\acmBooktitle{Proceedings of the 32nd ACM International Conference on Information and Knowledge Management (CIKM '23), October 21--25, 2023, Birmingham, United Kingdom}
\acmPrice{15.00}
\acmDOI{10.1145/3583780.3615236}
\acmISBN{979-8-4007-0124-5/23/10}

\usepackage[utf8]{inputenc} 
\usepackage[T1]{fontenc}    
\usepackage{hyperref}       
\usepackage{url}            
\usepackage{booktabs}       
\usepackage{amsfonts}       
\usepackage{nicefrac}       
\usepackage{microtype}      
\usepackage{xcolor}         
\usepackage{booktabs}
\usepackage{multirow}
\usepackage{tabularx}
\usepackage{todonotes}
\usepackage{amsmath,amsfonts}
\usepackage{algpseudocode}
\usepackage{algorithm}
\usepackage{graphicx}
\usepackage{textcomp}
\usepackage{xcolor}
\usepackage{csquotes}
\usepackage{float}
 \usepackage{balance}
\usepackage[normalem]{ulem}

\usepackage{fancyhdr}
\AtBeginDocument{%
    \addtolength{\footskip}{2.0\baselineskip}%
    \fancyfoot[L]{ \footnotesize
    © 2023 Copyright held by the owner/author(s). This is the author's version of the work. It is posted here for your personal use. Not for redistribution. The definitive Version of Record was published in Proceedings of the 32nd {ACM} International Conference on Information and Knowledge Management, {CIKM} 2023, Birmingham, United Kingdom, October 21-25, 2023, \url{http://dx.doi.org/10.1145/3583780.3615236}
    }%
}

\author{Anton Lee}
\orcid{0009-0008-6566-7785}
\affiliation{%
   \institution{Victoria University of Wellington}
   \city{Wellington}
   \country{New Zealand}
}
\additionalaffiliation{%
   \institution{University of Waikato}
   \city{Hamilton}
   \country{New Zealand}
}
\email{anton.lee@vuw.ac.nz}

\author{Yaqian Zhang}
\orcid{0000-0002-8594-4697}
\affiliation{%
   \institution{University of Waikato}
   \city{Hamilton}
   \country{New Zealand}}
\email{yaqian.zhang@waikato.ac.nz}

\author{Heitor Murilo Gomes}
\orcid{0000-0002-5276-637X}
\affiliation{%
   \institution{Victoria University of Wellington}
   \city{Wellington}
   \country{New Zealand}}
\email{heitor.gomes@vuw.ac.nz}

\author{Albert Bifet}
\orcid{0000-0002-8339-7773}
\affiliation{%
   \institution{University of Waikato }
   \city{Hamilton}
   \country{New Zealand}}
\additionalaffiliation{%
   \department{LTCI, Telecom Paris }
   \institution{Institute Polytechnique de Paris}
   \city{Paris}
   \country{France}}
\email{abifet@waikato.ac.nz}

\author{Bernhard Pfahringer}
\orcid{0000-0002-3732-5787}
\affiliation{%
   \institution{University of Waikato}
   \city{Hamilton}
   \country{New Zealand}}
\email{bernhard@waikato.ac.nz}

\begin{abstract}
Continual learning aims to create artificial neural networks capable of accumulating knowledge and skills through incremental training on a sequence of tasks. The main challenge of continual learning is catastrophic interference, wherein new knowledge overrides or interferes with past knowledge, leading to forgetting. An associated issue is the problem of learning "cross-task knowledge," where models fail to acquire and retain knowledge that helps differentiate classes across task boundaries. A common solution to both problems is "replay," where a limited buffer of past instances is utilized to learn cross-task knowledge and mitigate catastrophic interference. However, a notable drawback of these methods is their tendency to overfit the limited replay buffer. In contrast, our proposed solution, SurpriseNet, addresses catastrophic interference by employing a parameter isolation method and learning cross-task knowledge using an auto-encoder inspired by anomaly detection. SurpriseNet is applicable to both structured and unstructured data, as it does not rely on image-specific inductive biases. We have conducted empirical experiments demonstrating the strengths of SurpriseNet on various traditional vision continual-learning benchmarks, as well as on structured data datasets. Source code made available at \url{https://doi.org/10.5281/zenodo.8247906} and \url{https://github.com/tachyonicClock/SurpriseNet-CIKM-23}
\end{abstract}



\begin{document}
\title{Look At Me, No Replay! \\SurpriseNet: Anomaly Detection Inspired Class Incremental Learning}
\maketitle

\section{Introduction}
\label{sec:introduction}

Learning, in the biological sense, is a continuous process of acquiring knowledge and skills, where ideas build upon each other over time. This acquired knowledge is adaptable and adjusts to changes in the world, such as a favorite restaurant relocating or the introduction of new words. Our world is dynamic and constantly changing. In comparison, deep learning is restricted to learning from data that has been homogenized through shuffling to conform to a stationary data distribution. \textbf{\textit{Continual learning}} aims to free neural networks from interpreting data as stationary and enables them to learn continuously from a potentially infinite and evolving stream of data~\cite{parisi2019}.

Continual learning encompasses various scenarios, such as \textbf{Task Incremental Learning (Task-IL)} and \textbf{Class Incremental Learning (Class-IL)} \citep{journals/corr/abs-1904-07734}. Continual learning problems typically involve a sequence of tasks, where the learner is trained sequentially on each task before being evaluated on all tasks collectively. When new knowledge is acquired through training, it degrades old representations, negatively impacting the network's performance on past tasks \citep{mccloskeyCatastrophicInterferenceConnectionist1989a, french1999}. Even small changes can have a catastrophic effect. This phenomenon is known as "catastrophic interference" or forgetting.

In the Task-IL scenario, the learner has access to task labels or IDs during both training and testing. However, in many real-world applications, assuming the availability of task labels is impractical. On the other hand, the Class-IL scenario removes the learner's knowledge of the task label during testing. Empirically, there exists a significant performance gap between the Task-IL and Class-IL scenarios. In Class-IL, the learner must differentiate between classes across different tasks, rather than just within a single task as in Task-IL. Researchers refer to this challenge as "cross-task knowledge," "inter-task confusion," or "inter-task class separation" \citep{kim2021, masana2022, journals/corr/abs-2211-02633}.

To effectively learn continually in the Class-IL scenario, reducing catastrophic interference and acquiring cross-task knowledge becomes essential \cite{journals/corr/abs-2211-02633}.

In a recent survey, \citet{masana2022} concluded that replay-free methods cannot compete with replay methods. The replay mechanism involves storing a subset of each task and continuously replaying instances to prevent forgetting and gain cross-task knowledge. Unfortunately, replay has some drawbacks. Firstly, replay methods tend to overfit on the replay buffer \cite{delange2021,masana2022}. Secondly, replaying data through a neural network is inefficient, as the replay instances consume compute resources. Consequently, finding a mechanism for continual learning without replay holds significant value.

\begin{figure*}
    \centering
    \includegraphics[width=0.75\textwidth]{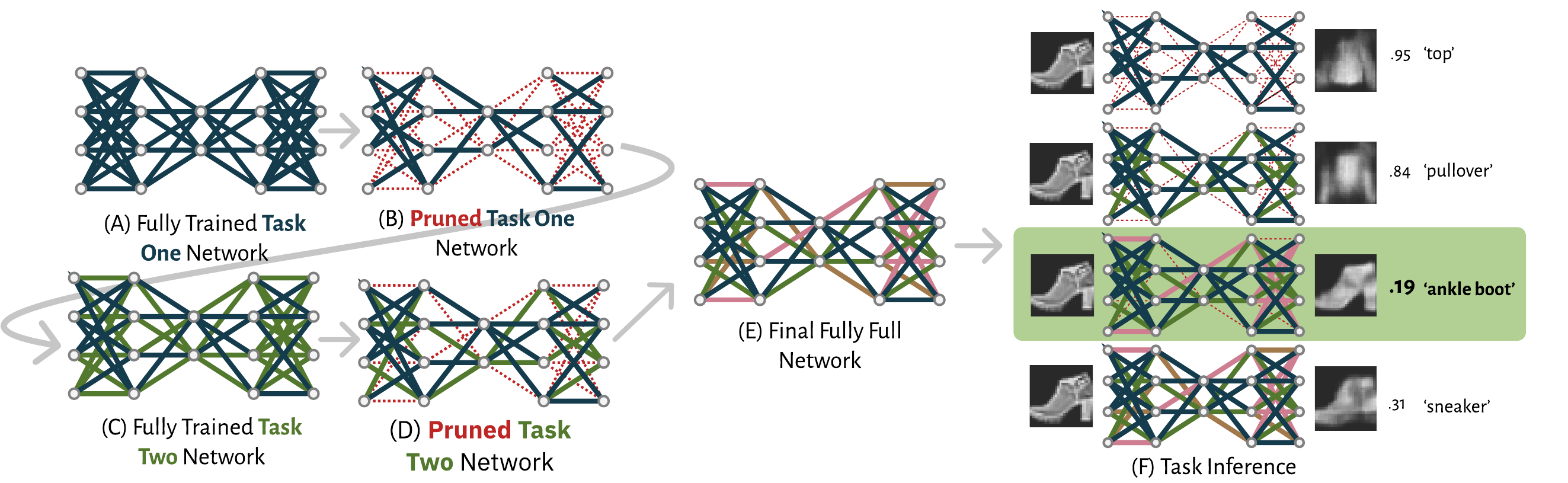}
    \caption{ 
    \textbf{(A)} Train a neural network on the classes in the first task.
    \textbf{(B/D)} \textit{Prune} the network by some proportion i.e. $\lambda=50\%$.
    \textbf{(C)} The blue weights are \textit{frozen} to avoid interference, but pruned weights are reused. 
    \textbf{(E)} ``Train, prune, retrain and freeze" repeats until tasks are exhausted. 
    \textbf{(F)} During evaluation, SurpriseNet infers the task by passing each instance through task-specific subsets in the network, generating multiple outputs. The task is identified by comparing the reconstruction quality.
}
    \label{fig:StepByStep}
\end{figure*}

In this work, we aim to bridge the gap between Class-IL and Task-IL by explicitly performing task identification with tools inspired by anomaly detection. Specifically, SurpriseNet employs an anomaly detection-inspired auto-encoder to infer tasks and applies parameter isolation to prevent catastrophic interference. It distinguishes itself from similar methods by being compatible with both structured and unstructured data, eliminating the requirement for image-specific inductive biases.

\section{Related Work}
\paragraph{Continual Learning and Anomaly Detection}
\citet{journals/corr/abs-2211-02633, conf/cvpr/KimEX022} demonstrated the connection between task identification and anomaly detection. Their work led to the proposal of CLOM (Continual Learning based on OOD detection and Task Masking), a state-of-the-art replay-free method that combines anomaly detection and fixed network parameter isolation~\cite{conf/cvpr/KimEX022,journals/corr/abs-2211-02633}. CLOM uses image rotations and other data augmentations to serve as out-of-distribution training data during contrastive learning. These assumptions make CLOM inappropriate for some vision datasets without modification because different problems require non-trivial data-specific decisions.

In addition to CLOM, we compare SurpriseNet against replay, pseudo-replay, and regularization strategies. We choose the replay strategies: DER \cite{conf/nips/BuzzegaBPAC20}, A-GEM \cite{conf/iclr/ChaudhryRRE19}, iCaRL \cite{conf/cvpr/RebuffiKSL17}, CLOM-c \cite{conf/cvpr/KimEX022}, and experience replay. Replay methods reuse a limited buffer of past training data to avoid interference and gain cross-task knowledge, but vary in how samples are chosen and exploited. Our experiments use a small replay buffer of 500 to emphasize sample efficiency~\cite{conf/nips/BuzzegaBPAC20}. Apart from normal replay, we compare against pseudo-replay strategies: BIR~\cite{van_de_ven2020} and GR~\cite{van_de_ven2020}. These methods generate training data to resemble past data and approximate replay. Finally, we compare against regularization strategies EWC~\cite{journals/corr/KirkpatrickPRVD16} and LwF~\cite{li2017}, which aim to slow changes to important weights in the network to avoid interference. These methods typically lack a specific mechanism to learn cross-task knowledge and, consequently, perform poorly on class-IL scenarios.

Parameter isolation techniques isolate and protect parameters important to previous tasks, keeping them fixed while allowing other parameters to be learnt for new tasks. ``Hard Attention to The Task" (HAT) protects past tasks by learning a mask controlled by a task label embedding \cite{conf/icml/SerraSMK18}. Similarly, "PackNet" protects past tasks by creating a mask using weight magnitude as a heuristic for importance \cite{mallya2018}. We do not compare against HAT or PackNet because they rely on a task ID to activate and deactivate the correct subsets of their neural networks.

In contrast to replay methods, SurpriseNet effectively avoids overfitting the replay buffer and enhances training efficiency by eliminating the need to replay samples. While regularization strategies are also replay-free, EWC and LwF encounter difficulties in the Class-IL setting. SurpriseNet shares the concept of fixed network parameter isolation with HAT and PackNet but is not limited to Task-IL since it can infer task IDs. This characteristic aligns SurpriseNet closely with CLOM in terms of theoretical similarity, but both approaches use different mechanisms for parameter isolation and anomaly detection. Our approach has an advantage over CLOM as it can be applied to a wider range of datasets and makes fewer assumptions about the data.

\section{Method}

SurpriseNet divides continual learning into two subproblems: preventing catastrophic forgetting and identifying task IDs. The issue of catastrophic forgetting is tackled by pruning un-important parameters and freezing the important ones, creating ``task-specific subsets" for each task. These subsets can be selectively activated or deactivated to solve each task in isolation. The second problem, task identification, is addressed with an auto-encoder. This auto-encoder gauges the reconstruction quality of each task-specific subset, to identify the task. This is based on the assumption that the most accurate reconstruction corresponds to the instance's task. Integrating these techniques produces an effective continual learner.

SurpriseNet builds upon PackNet~\cite{mallya2018}, inheriting the pruning procedure and parameter importance heuristic. However, it incorporates additional features such as task inference, a new pruning schedule, and a hybrid architecture. Figure \ref{fig:StepByStep}(A-E) provides a summary of the PackNet procedure as utilized in SurpriseNet. Most importantly, unlike PackNet, SurpriseNet is not confined to Task-IL as it is capable of inferring task IDs.

SurpriseNet introduces a novel pruning schedule (\textbf{EqPrune}), which allocates an equal proportion of the network to each task. Our experiments reveal that EqPrune consistently performs well when compared to pruning with a sub-optimal $\lambda$, although it is often outperformed by a well-chosen $\lambda$. EqPrune is a viable default when, hyper-parameter tuning is not possible.

\label{sec:task_identification}
Unlike PackNet, SurpriseNet is a hybrid supervised and unsupervised learner. It consists of an encoder $z = E_\theta(x)$, a decoder $\hat x = D_\theta(z)$, and a classifier $\hat y = C_\theta(z)$. The encoder and decoder form an auto-encoder (AE) where $\hat x = D_\theta(E_\theta(x))$. The classifier connects to the encoder, $\hat y = C_\theta(E_\theta(x))$. SurpriseNet is trained to achieve dual objectives. The first objective reconstructs the input in the output using mean squared error as the loss function $L_{rec} = \frac{1}{n} \sum_{i=1}^n ||x_i - D_\theta(E_\theta(x_i))||^2$. The second objective classifies instances with cross-entropy loss $L_{cls} = -\frac{1}{n} \sum_{i=1}^n y_i \log(C_\theta(E_\theta(x_i)))$. Here, $(x_i, y_i)$ represent an instance in a batch of size $n$ originating from the active task. Adding the losses together forms a hybrid model outputting both reconstructions and a classification at each forward step. As an alternative to AE, we experimented with variational auto-encoders (VAE) \cite{kingma2013}, which regularize the latent space to follow a probability distribution. The intention was to learn a more structured internal representation.

The auto-encoder infers the task, using a technique inspired by anomaly detection. The auto-encoder model contains a bottleneck that must learn a compressed representation, containing regularities of nominal training data. Consequently, anomalous data belonging to other tasks, is harder to reconstruct \cite{pang2022}. Essentially, the best-suited task-specific subset can be found by measuring the lowest reconstruction loss (e.g $||x-\hat x||^2$) of each task-specific subset. Figure \ref{fig:StepByStep} (E-F) further describes the SurpriseNet task inference procedure.


\label{sec:SurpriseNetE}
SurpriseNet's ability to perform anomaly detection and infer tasks depends on the capabilities of AE and VAE. However, anomaly detection remains a challenging problem, particularly for high-dimensional data such as images. In certain cases, deep generative models assign higher likelihood to anomalous data i.e. lower reconstruction loss~\cite{nalisnick2018}. This has been attributed to the emphasis of generative modeling on low-level features~\cite{conf/icml/HavtornFHM21}. To overcome this limitation, we utilize a pre-trained ResNet18 to extract features and reduce dimensionality, enabling our model to work with higher-level features. \citet{van_de_ven2020} argues that the outer regions of the human mind, which undergo minimal changes throughout life, perform compression and dimensionality reduction, providing a biological basis for a similar procedure. Our approach, incorporating a pre-trained ResNet18, is designated \textbf{SurpriseNetE}.

\begin{table*}
    \caption{Mean Final Accuracy (\%) $\pm$ One Standard Deviation, from 5 runs}
    \begin{center}
        \begin{tabularx}{\textwidth}{Xllllll}
            \toprule
            Strategy &  & S-DSADS & S-PAMAP2 & S-FMNIST & S-CIFAR10 & S-CIFAR100 \\
            \midrule
             &  & \footnotesize{(2 classes for 9 tasks)} & \footnotesize{(2 classes for 6 tasks)} & \footnotesize{(2 classes for 5 tasks)} & \footnotesize{(2 classes for 5 tasks)} & \footnotesize{(10 classes for 10 tasks)} \\
            Joint (Non-CL) &  & 75.08 $\pm$ 3.33 & 88.12 $\pm$ 1.52 & 87.86 $\pm$ 0.72 & 90.35 $\pm$ 0.83 & 72.62 $\pm$ 0.58 \\
            \midrule
            \multicolumn{7}{c}{\textit{Replay Method $n$ = Buffer Size}} \\
            Experience Replay & n=500 & 74.40 $\pm$ 3.29 & \textbf{81.30 $\pm$ 4.42} & 76.08 $\pm$ 2.07 & 44.24 $\pm$ 7.51 & 21.21 $\pm$ 0.87 \\
            A-GEM \cite{conf/iclr/ChaudhryRRE19} & n=500 & 29.34 $\pm$ 6.83 & 48.41 $\pm$ 11.21 & 53.27 $\pm$ 7.86 & 24.05 $\pm$ 5.30 & 9.58 $\pm$ 0.55 \\
            DER \cite{conf/nips/BuzzegaBPAC20} & n=500 & 65.02 $\pm$ 5.30 & 69.86 $\pm$ 9.19 & 80.71 $\pm$ 0.59 & 57.41 $\pm$ 4.46 & 25.18 $\pm$ 7.35 \\
            iCaRL \cite{conf/cvpr/RebuffiKSL17} & n=500 & 43.74 $\pm$ 17.97 & 71.18 $\pm$ 2.29 & 57.22 $\pm$ 17.64 & 53.72 $\pm$ 7.00 & 41.05 $\pm$ 0.46 \\
            CLOM-c \cite{conf/cvpr/KimEX022} & n=(20 per class) & NA & NA & 37.88 $\pm$ 9.19 & \textbf{81.37 $\pm$ 0.52} & \textbf{71.90 $\pm$ 1.53} \\
            \multicolumn{7}{c}{\textit{Replay Free}} \\
            Naive SGD &  & 9.81 $\pm$ 2.15 & 15.02 $\pm$ 2.85 & 19.85 $\pm$ 0.33 & 18.53 $\pm$ 1.36 & 8.80 $\pm$ 0.39 \\
            BIR \cite{van_de_ven2020} &  & 14.68 $\pm$ 4.08 & 78.01 $\pm$ 1.65 & 76.37 $\pm$ 1.87 & NA & 21.2 $\pm$ 1.06 \\
            LwF \cite{li2017} &  & 10.77 $\pm$ 0.86 & 16.38 $\pm$ 0.08 & 20.07 $\pm$ 0.54 & 18.85 $\pm$ 0.30 & 8.80 $\pm$ 0.39 \\
            EWC \cite{journals/corr/KirkpatrickPRVD16} &  & 11.13 $\pm$ 0.43 & 16.52 $\pm$ 0.11 & 22.27 $\pm$ 5.55 & 18.60 $\pm$ 1.09 & 7.90 $\pm$ 0.72 \\
            GR \cite{van_de_ven2020} &  & 13.02 $\pm$ 2.68 & 63.40 $\pm$ 4.04 & 76.37 $\pm$ 1.87 & 18.93 $\pm$ 1.06 & 6.92 $\pm$ 0.54 \\
            CLOM \cite{conf/cvpr/KimEX022} &  & NA & NA & 54.51 $\pm$ 4.04 & 79.78 $\pm$ 1.16 & 71.50 $\pm$ 1.61 \\
            (ours) SurpriseNet & VAE & 76.95 $\pm$ 1.08 & 79.76 $\pm$ 1.61 & 79.20 $\pm$ 1.72 & 37.21 $\pm$ 1.66 & 15.85 $\pm$ 0.28 \\
            (ours) SurpriseNet & AE & \textbf{78.21 $\pm$ 3.83} & 80.48 $\pm$ 1.59 & \textbf{82.16 $\pm$ 0.39} & 38.29 $\pm$ 2.01 & 16.49 $\pm$ 0.58 \\
            (ours) SurpriseNet & VAE EqPrune & 77.56 $\pm$ 1.22 & 79.45 $\pm$ 1.93 & 79.47 $\pm$ 1.95 & 34.86 $\pm$ 2.00 & 14.31 $\pm$ 0.54 \\
            (ours) SurpriseNetE & VAE EqPrune & NA & NA & 80.52 $\pm$ 2.06 & 76.04 $\pm$ 1.42 & 48.15 $\pm$ 0.68 \\
            \bottomrule
        \end{tabularx}
    \end{center}
    \label{tab:comparison}
\end{table*}
\begin{figure*}
    \centering
    \includegraphics[width=\textwidth]{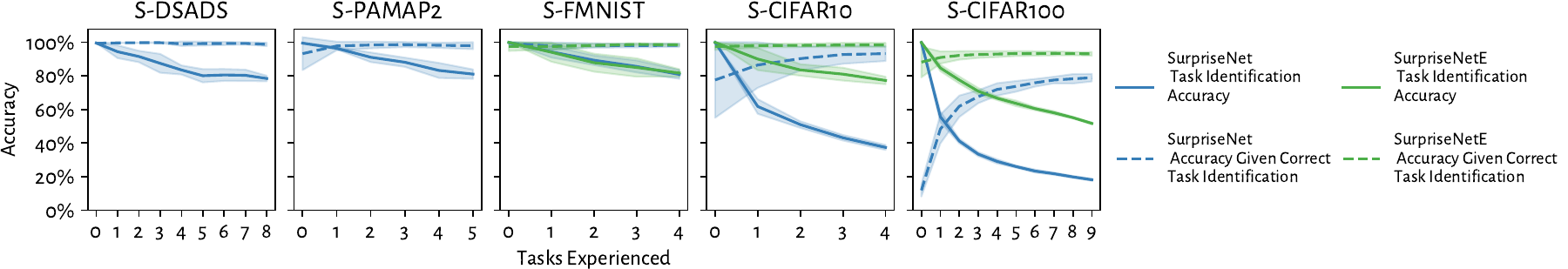}
    \caption{To classify correctly SurpriseNet must first identify a task, then classify within the task. ``SurpriseNet VAE EqPrune" and ``SurpriseNetE VAE EqPrune" configurations are shown above where each approach is applicable.}
    \label{fig:SurpriseNetAccOverTime}
\end{figure*}

\subsection{Experimental Design}
To improve the comparison with other methods, we adhere to the desirable qualities of a continual learning experimental setting outlined by \citet{farquhar2019}. These qualities include cross-task resemblances, a shared output head, no assumption of task labels at test time, no unconstrained retraining on old tasks, and scenarios involving more than two tasks. Furthermore, we randomly sample the task order to showcase the robustness of our method to task ordering. 

SurpriseNet is demonstrated on vision datasets with classes split into multiple tasks: \textbf{S-FMNIST} (epochs=20) \cite{xiao2017}, \textbf{S-CIFAR10} (epochs=50) \cite{krizhevsky2009}, and \textbf{S-CIFAR100} (epochs=100) \cite{krizhevsky2009}. S-CIFAR100 is particularly popular among researchers \cite{van_de_ven2020, delange2021, journals/nn/MundtHPR23, kim2021}.

In addition to vision continual learning benchmarks, SurpriseNet is benchmarked on structured data datasets, specifically human activity recognition datasets (HAR). HAR is a desirable domain for continual learning because new unique activities will eventually occur~\cite{journals/tecs/LeiteX22}. \citet{journals/isci/JhaSZY21} provides the example of COVID-19, the outbreak significantly changed human behavior, something a continual learning system could adapt to without also forgetting normal behavior. \textbf{S-DSADS} (Epoch Budget=200) \cite{conf/iswc/ReissS12} splits the Daily and Sports Activities dataset of sensor data for 19 activities randomly into 9 tasks of 2 activities each (the remaining activity is ignored). Participants 7 and 8 are used in the test set. \textbf{S-PAMAP2} (Epoch Budget=200) \cite{journals/prl/ChavarriagaSCDTMR13} splits the Physical Activity Monitoring dataset, containing data of 12 different physical activities, into 6 tasks of 2 activities each. Participants 7 and 8 are used in the test set. The variant of the dataset we utilize is provided alongside our code and uses a sliding window to generate features from raw accelerometer data for both S-DSADS and S-PAMAP2~\cite{journals/tiot/YeNSJZ21, journals/isci/JhaSZY21, journals/corr/abs-2007-03032, conf/percom/WangCHPY18}.

To fairly represent each method hyper-parameters must be optimized. Grid searches were conducted for EWC, LWF, ICARL, AGEM, DER, and ER, exploring learning rates and additional strategy-specific hyper-parameters where applicable. CLOM, BIR, and GR, use their default configurations (applying MNIST's for FMNIST), but epochs were reduced to fall within our epoch budget. Minor modifications were made to BIR and GR to ensure compatibility with S-DSADS and S-PAMAP2. Grid searches were also performed to select optimal values for fc-units, latent-dims, and learning-rate for both BIR and GR on those datasets. Regarding SurpriseNet, a grid search was conducted to determine the best prune-proportion, while keeping the learning rates constant. Specifically, S-FMNIST, S-CIFAR10, and S-CIFAR100 employed Adam with a learning rate of 0.0001, while S-DSADS and S-PAMAP2 used Adam with a learning rate of 0.0008. SurpriseNet is a ResNet like auto-encoder for vision datasets and a multilayer perceptron auto-encoder on non-vision datasets. In vision datasets using SurpriseNetE a feature extractor feeds into an auto-encoder multilayer perceptron. Results from grid search and hyper-parameter selections can be found alongside the source code.

\section{Results}
\label{sec:results}
Table \ref{tab:comparison} compares SurpriseNet with other strategies. SurpriseNet proves to be effective, surpassing replay methods (n=500) and other replay-free approaches when applied to lower-dimensional datasets such as S-DSADS, S-PAMAP2, and S-FMNIST. However, SurpriseNet's effectiveness diminishes when dealing with higher-dimensional data, as observed in S-CIFAR10 and S-CIFAR100. This limitation can be attributed to the fact that deep generative models are not well-suited for anomaly detection tasks involving high-dimensional data \cite{nalisnick2018}.

To address this issue, SurpriseNetE use a pre-trained network to reduce dimensionality. Figure \ref{fig:SurpriseNetAccOverTime} demonstrates the improvement in task identification. SurpriseNetE continues to outperform most other methods in our experiments, with the exception of CLOM, given our buffer budget and epoch budget.

The variances observed in Table \ref{tab:comparison} exceed those reported by other researchers in similar experiments due to our experimental protocol's inclusion of class order shuffling. To illustrate the effect of this factor, we repeated our iCaRL experiments on S-FMNIST without shuffling class ordering. The sample's standard deviation decreased significantly from $57.2\pm17.6$ to $65.3\pm1.2$. We strongly argue that shuffling class orders is necessary to evaluate a model's robustness against task composition and to avoid biases in default class orderings.

\section{Conclusion}
SurpriseNet utilizes parameter isolation to stabilize against forgetting and an auto-encoder to learn cross-task knowledge and identify tasks. High-dimensional data challenges our method but we explore a dimensionality reduction technique to mitigate the limitation. Future work could focus on improving the pruning procedure to enable more efficient utilization of capacity or enhancing the anomaly detection capabilities. Although most continual learning works focus on image data, we present a more flexible method. SurpriseNet is a flexible, effective, and replay-free method competitive against replay methods, in some scenarios.

\newpage
\balance
\bibliography{bib}

\end{document}